# Tokensome: Towards a Genetic Vision-Language GPT for Explainable and Cognitive Karyotyping

Haoxi Zhang [1], Yuanxin Lin [1*], Xinxu Zhang [1*], Maiqi Wang [1*], Yi Lai [2], Yu Wang [3*], Linfeng Yu [1*], Yufeng Xu [1*], Ran Cheng [1*], Edward Szczerbicki [4]

**Abstract:** Artificial Intelligence (AI) has become a transformative force in medical diagnostics, notably automating karyotyping, a key aspect of genetic analysis. However, the problem of automatic karyotype analysis is often defined as a visual perception task focused solely on chromosomal object-level modeling. This definition has led most existing methods to overlook componential and holistic information, significantly constraining model performance. Moreover, the lack of interpretability in current technologies hinders clinical adoption. In this paper, we introduce Tokensome, a novel vision-language model based on chromosome tokenization for explainable and cognitive karyotyping. This approach tokenizes chromosomes at the sub-chromosome level, representing these subdivisions as semantically meaningful image segments derived from self-supervised learning vision models, and feeds them into a Large Language Model (LLM) to learn a vision-language model for chromosome. Tokensome elevates the method from the conventional visual perception layer to the cognitive decision-making layer. This elevation enables the integration of domain knowledge and cognitive reasoning via knowledge graphs and LLMs, markedly enhancing model's explainability and facilitating abnormality detection. In addition, by leveraging domain knowledge, such as the anticipated chromosome count per image, Tokensome enhances karyotyping performance through a holistic, strategy-driven analysis over the entire image, avoiding the limitations of object-focused and perception-based analysis. This advancement reduces ambiguities and inconsistencies, ensuring domain-aligned optimizations. To the best of our knowledge, Tokensome is the first model that successfully performs integrated segmentation, classification, and abnormality detection like human experts. Tokensome pioneers a trustworthy approach for AI-assisted karyotyping, offering an explainable and cognitive solution that harmonizes predictive accuracy with transparency. Our approach demonstrates that embedding higher-order cognitive capabilities into conventional vision models can revolutionize biomedical image analysis and has broad implications for enhancing healthcare diagnostics.

# Contents

## 1. Introduction

Artificial intelligence (AI) has achieved significant progress due to recent rapid advances in deep learning [1] techniques. Through deep learning's powerful automated feature extraction capacities—enabling detection of nuanced multidimensional patterns in medical images beyond human discernment—AI holds considerable potential to unlock substantial improvements in medical imaging domain [2][3]. However, integration of AI technologies into real-world clinical settings remains severely limited. A major obstacle is the predominant opacity of state-of-the-art AI systems, which frequently manifest as inscrutable "black-box" models that provide no actionable evidence or confidence metrics to justify their output decisions and predictions [4]. This lack of model explainability or interpretability not only hinders scientific understanding of system behavior for imaging analysis tasks, but also critically erodes clinician and patient trust.

Karyotyping is a vital cytogenetic task for detecting genetic abnormalities by analyzing metaphase chromosome images [5]. The process involves first preparing a complete set of micro-photographed metaphase chromosomes from cell samples. This includes properly segmenting, classifying, and pairing the 23 chromosome types into homologous pairs to produce a karyogram (see Figure 1). Subsequently, the karyogram is then carefully analyzed by clinical cytogeneticists to identify any anomalies [6][7].

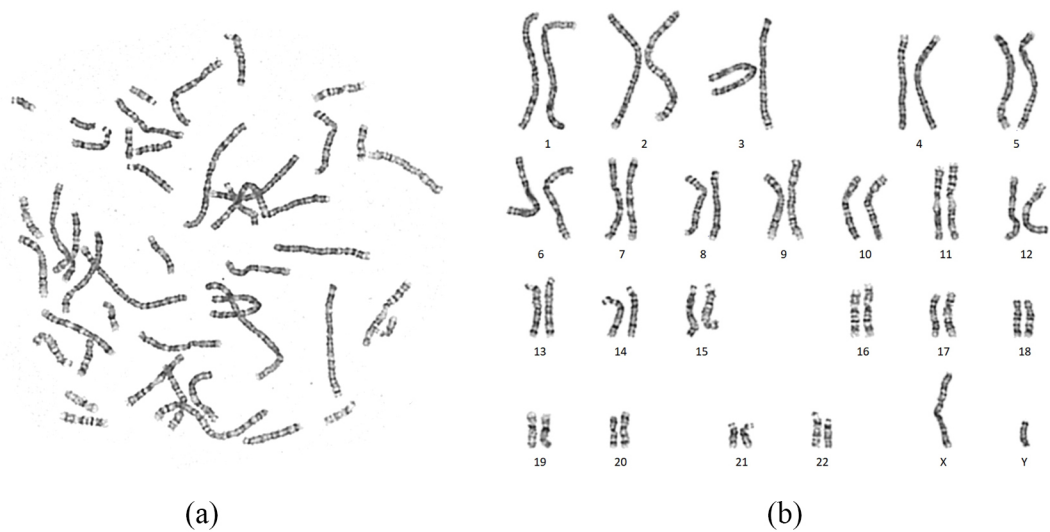

(a)                  (b)

**Figure 1:** (a) A G-stained microscopic image of male chromosomes for one case. (b) The karyogram of (a).

The task is challenging due to the highly condensed, coiled, and sometimes overlapped nature of metaphase chromosomes, making the process labor-intensive and time-consuming even for experienced cytogeneticists. To alleviate this burden, computer-assisted chromosome analysis and automated karyotyping have been longstanding endeavors in the field, with dedicated efforts from numerous researchers over recent decades [3][8][9]. Despite these advances, current tools and efforts tend to only tackle individual tasks such as chromosome segmentation or classification. This fragmented approach lacks the comprehensive, end-to-end system necessary for fully integrated analysis. Crucially, these tools also fall short in explainability, an essential component for clinical confidence and understanding. Furthermore, the problem of automatic karyotype analysis is often defined as a visual perception task focused solely on chromosomal object-level modeling. This definition has led most existing methods to overlook componential and holistic information, significantly constraining model performance.

In this paper, we introduce Tokensome, a novel vision-language model based on chromosome tokenization for explainable and cognitive karyotyping. This approach tokenizes chromosomes at the sub-chromosome level, representing these subdivisions as semantically meaningful image segments derived from self-supervised learning vision models, and feeds them into a Large Language Model (LLM) to learn a vision-language model for chromosome. Tokensome elevates the method from the conventional visual perception layer to the cognitive decision-making layer. This elevation enables the integration of domain knowledge and cognitive reasoning via knowledge graph and LLM, markedly enhancing model's explainability and facilitating abnormality detection. In addition, by leveraging domain knowledge, such as the anticipated chromosome count per image, Tokensome enhances karyotyping performance through a holistic, strategy-driven analysis over the entire image, avoiding the limitations of object-focused and perception-based analysis. This advancement reduces ambiguities and inconsistencies, enabling Tokensome to offer: (1) higher-order cognitive capability for karyotyping; (2) explainable and integrated chromosome instance segmentation, classification, and abnormality detection; (3) interactive online learning and domain expertise incorporation; and (4) active leveraging of domain knowledge for image-wide strategic reasoning and domain-aligned optimization.

## 2. Method

The core of our method is the chromosome tokenization, which processes chromosomes at the sub-chromosome level. It represents these subdivisions as semantically meaningful image segments derived from self-supervised vision models, and integrates them into a Large Language Model (LLM) to develop a vision-language model for chromosome analysis. This process consists of three steps: sub-chromosome representation learning, positional encoding, and vision-language modeling.

### 2.1 Sub-Chromosome Representation Learning

Sub-chromosomes represent semantically meaningful segments of chromosomes at the sub-object level. The initial phase of the Tokensome process involves a comprehensive analysis of chromosomal images and extraction of intricate features to form these sub-chromosomes. Inspired by the emerged properties in DINO [10], which contain explicit information about the semantic segmentation of an image, we utilize DINO for initial sub-chromosome representation learning. As shown in Figure 2, we input various crops into both the student and teacher networks to generate patch-level cluster predictions and determine optimal cluster assignment targets. This requires an alignment for cluster targets and assignments. Moreover, by utilizing the ViT's attention map, we enhance clustering precision on foreground features.

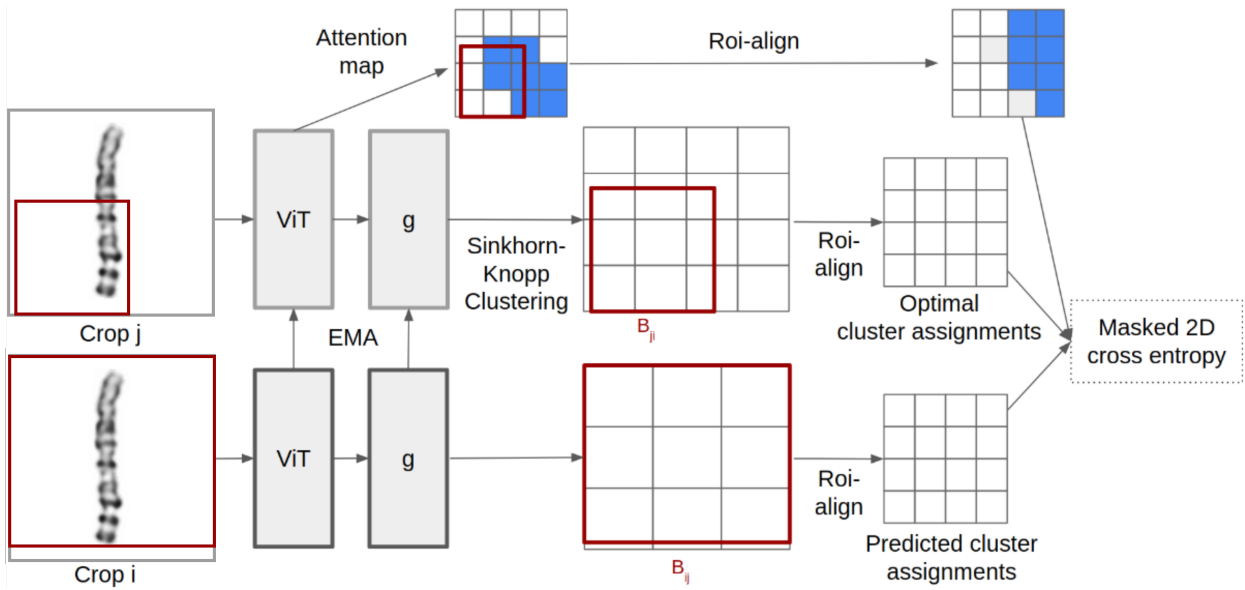

**Figure 2:** The training pipeline of Tokensome.

However, the inherent complexity and challenges of chromosome images could constrain the comprehensiveness of sub-chromosomes identified by DINO. To address this issue, we implement post-processing using conditional random fields (CRF) [11] and overclustering [12] techniques to refine the sub-chromosome masks. Through this post-processing, we achieve fine and robust sub-chromosome representation for each chromosome type (see Figure. 3).

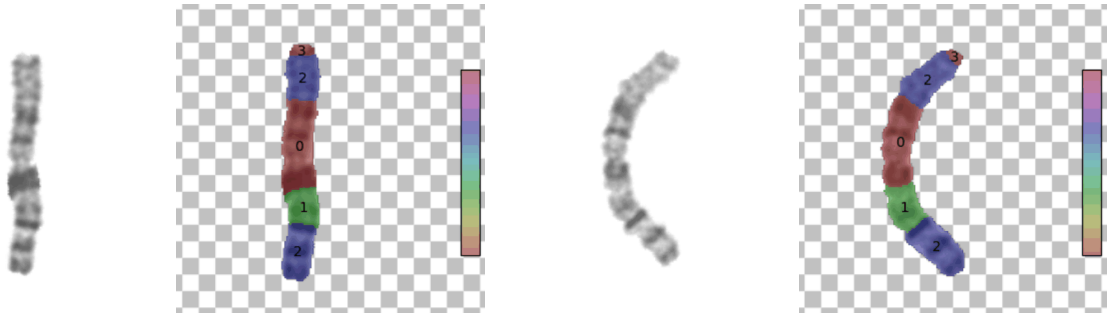

**Figure 3:** The learned sub-chromosomes.

## 2.2 Positional Encoding

It is crucial to clarify the spatial positional relationships between sub-chromosomes for accurate chromosome modeling. Given that chromosomes are fundamentally sequential, we employ the longitudinal axis to ascertain the specific spatial arrangement of sub-chromosomes. We have devised an erosion-based method to derive the longitudinal axis for each chromosome, effectively accommodating chromosomes of varied shapes (see Figure 4).

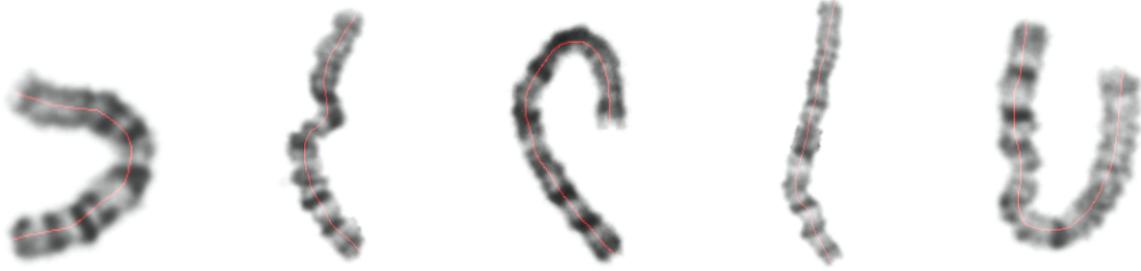

**Figure 4:** The chromosome longitudinal axis generation (depicted as the red lines).

By analyzing the coordinates of the longitudinal axis, we establish the start and end positions of the chromosome, and consequently, the sequence of its sub-chromosomes. Since the final representations of various sub-chromosomes have uniform dimensions, positional encoding can be directly applied to the sub-chromosome representations, akin to the process in LLMs.

**2.3 Vision-Language Modeling**

Having clustered the sub-chromosomes and determined their spatial positions for each chromosome type, we can directly derive the sequence patterns of each type using sequence mining algorithms such as FP-Growth [13].

However, conventional algorithms fall short in handling the variability of features and discerning subtle nuances critical for our model. To establish an end-to-end, robust, and flexible framework, we conceptualize the chromosome sequence as a unique genetic language, where each sub-chromosome serves as a 'token' within this linguistic framework, facilitating the processing of extensive end-to-end learning seamlessly. Drawing inspiration from Chen et al. [14], we adopt a simple yet effective approach for vision-language modeling and introduce a pair of special tokens, <SOC> and <EOC>, to denote the start and end of sub-chromosome tokens for each chromosome, respectively. For training, we employ the Phi-2 model [15], optimizing our system for comprehensive learning and analysis.

## 3. Experiment

### 3.1 Classification

**A. Datasets**

To evaluate the performance of our proposed method, we use the publicly available dataset reported in the recent study [16] named as *Dataset A* and one private dataset created with the West China Second Hospital named as *Dataset B*.

For *Dataset A* there are only 65 cases total (32 male and 33 female karyotypes). Usually, one case comprises 46 individual chromosomes, which means 2990 total chromosome samples for 24 classes. Specifically, for 22 autosomes (labelled as 0 to 21), there are 130 samples per class, while for sex chromosomes, there are 98 (33 × 2 + 32) X samples labelled as 22 and only 32 Y samples labelled as 23.

Correspondingly, there are 580 cases for *Dataset B*. The full statistics for both datasets are presented in Table 1. We split each dataset into training and testing sets using the ratio of 90:10.

**Table 1.** Statistics of the datasets.

| Dataset | Case | | Image per class | | | Total image |
|---|---|---|---|---|---|---|
| | Male | Female | 0 ~ 21 | 22 | 23 | |
| A | 32 | 33 | 130×22 | 98 | 32 | 2,990 |
| B | 256 | 324 | 1160×22 | 904 | 256 | 26,680 |

**B. Results**

To examine our proposed method in comparison with other approaches, we selected three primary CNN networks—AlexNet [17], VGG-16, and ResNet-50—as baseline methods, as well as three advanced models reported in [16] to serve as benchmarks against our proposed method on *Dataset A*. Moreover, we select five state-of-the-art methods published in the last five years, as summarized in Table 2.

The comparison is organized into two categories: *Category One*, which comprises the first 7 rows and is based on *Dataset A*; and *Category Two*, which includes the last 6 rows and is based on each method's respective dataset (our method is based on *Dataset B* in this category).

To perform the classification task, we use the backbone trained in sub-chromosome representation learning, and add a linear classifier at the end. The experimental results reveal that our proposed approach outperforms the other methods in *Category One*, achieving a classification accuracy of 98.96%. It is noteworthy that CIR-Net, to the best of our knowledge, represents the state-of-the-art method on this public dataset [16]. In *Category Two*, the five recent methods used for comparison are very strong benchmarks for evaluating classification performance. Our proposed method employs *the second smallest dataset* of 26,680 images—a count that is still about 10 times fewer than what KaryoNet uses. Despite this smaller dataset, our method achieves the highest accuracy among all compared methods across both categories.

**Table 2.** Performances of our method compared with state-of-the-art methods.

| Method | Accuracy | Dataset | |
|---|---|---|---|
| | | Total Image | Public |
| AlexNet | 83.86 | 2,990 | Yes |
| VGG-16 | 91.22 | 2,990 | Yes |
| ResNet-50 | 91.92 | 2,990 | Yes |
| Vanilla-CNN | 86.44 | 2,990 | Yes |
| SiameseNet | 87.63 | 2,990 | Yes |
| CIR-Net | 95.98 | 2,990 | Yes |
| ***Ours*** | **98.96** | 2,990 | Yes |
| Varifocal-Net [18] (2019) | 98.7 | 87,831 | NO |
| CIR-Net [16] (2020) | 95.98 | 2,990 | Yes |
| Zhang *et al*. [19] (2021) | 98.1 | 32,810 | NO |
| ETPC [20] (2022) | 95.3 | 28,225 | NO |
| KaryoNet [21] (2023) | 99.58 | 291,603 | NO |
| ***Ours*** | **99.71** | 26,680 | NO |

## 3.2 Abnormality Detection

Chromosome abnormalities fall into two main categories: numerical and structural anomalies. Numerical anomalies can be identified through instance counting, which is relatively straightforward. However, detecting structural abnormalities requires precise recognition of changes in chromosome structure, such as inversions, deletions, and others, which is very challenging. Consequently, there are fewer research findings in this area, and most studies focus on simple cases. For example, the study by Yan et al. [22] focused solely on identifying a translocation abnormality t(9;22) related only to chromosomes 9 and 22. Similarly, methods by Cox et al. [23] and Yang et al. [24] can only recognize some common structural abnormalities. In the research on general structural abnormality detection, a more recent work, UC-Det [25], employs techniques like the Squeeze and Excitation module and Convolutional Spatial Pooling Attention block, achieving an accuracy rate of over 99% in counting tasks. However, its accuracy for detecting structural abnormalities is only 75%.

We utilize the sub-chromosome representation with position encoding for structural abnormality detection tasks, conducting preliminary experiments on our abnormal chromosome dataset, which comprises 1,144 chromosomes, evenly split between normal and abnormal specimens.

The results are displayed in Figure 5: each point in the figure represents one experiment, with each experiment corresponding to a false negative rate and false positive rate. Using the Pareto optimal solution curve [26], we selected viable points (red dots) for structural abnormality detection. In the initial experiments, our model achieves a false negative rate of 0.016 (1.6%) and a false positive rate of 0.098 (9.8%).

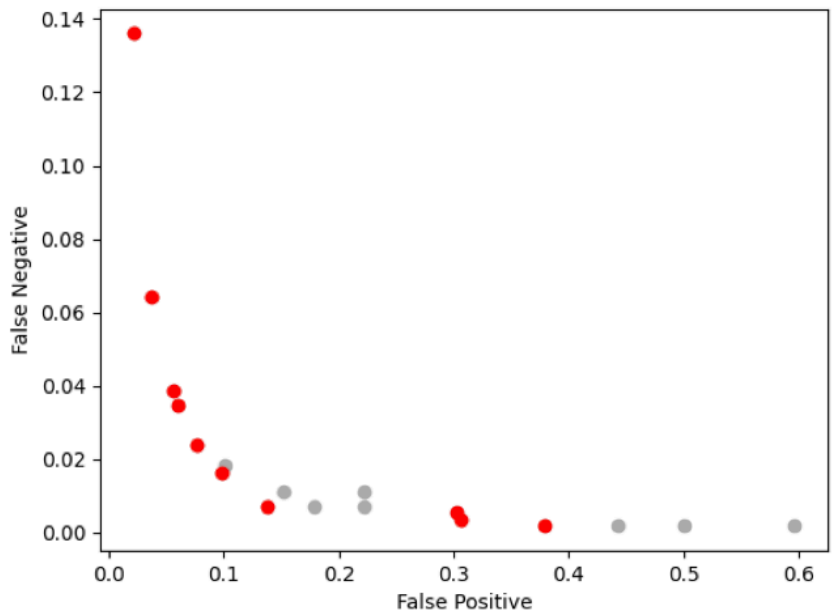

**Figure 5:** Structural abnormality results under different model parameters.

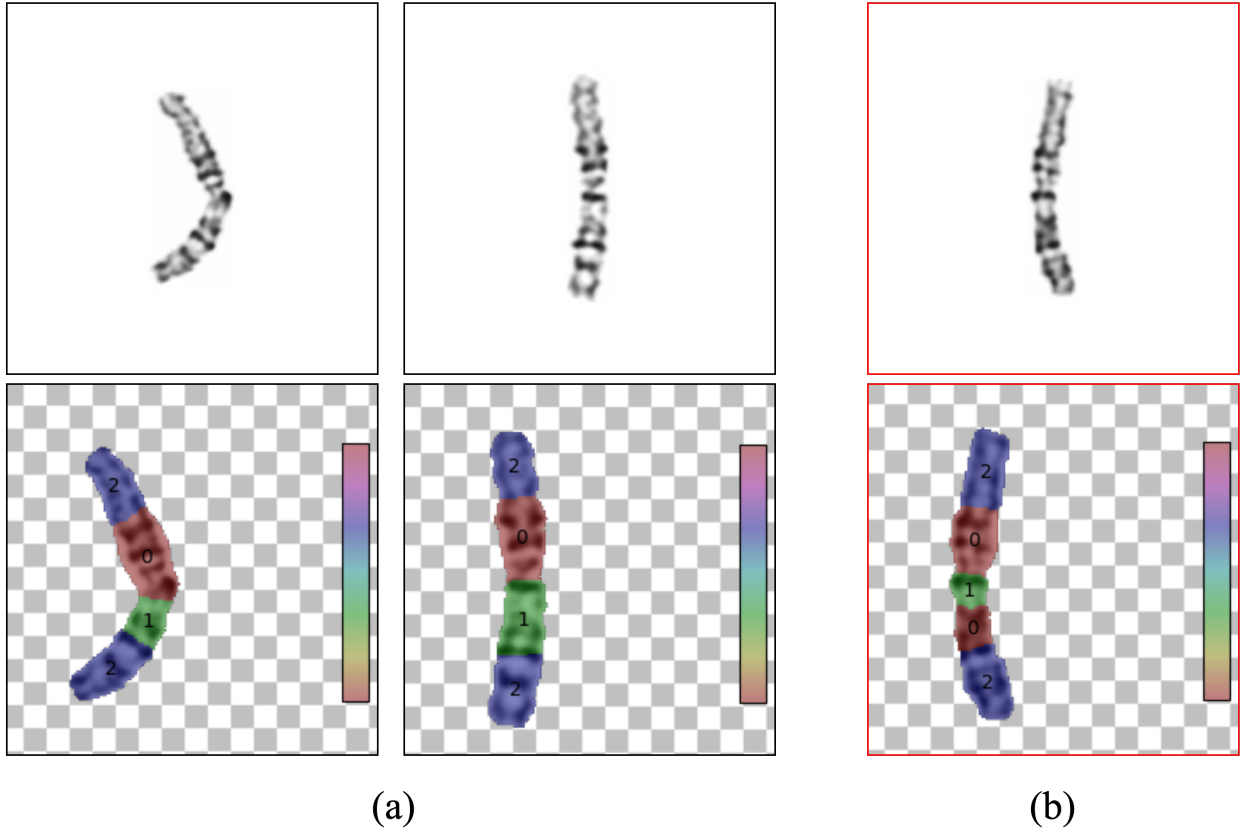

**Figure 6:** Explainable abnormality detection using sub-chromosomes: (a) Normal; (b) Abnormal

### 3.3 Explainability

With our sub-chromosome representations, Tokensome enhances explainability in segmentation, classification, and abnormality detection tasks. For instance, in our preliminary representation learning model, type 1 chromosome shows a sub-chromosome sequence pattern of "2-0-1-2," as illustrated in Figure 6 (a). An abnormal type 1 chromosome from our abnormal chromosome dataset was randomly chosen and analyzed by Tokensome, revealing a distinct sub-chromosome sequence pattern of "2-0-1-*0*-2," depicted in Figure 6 (b). This pattern diverges from the normal sequence of type 1 chromosome. Karyotype analysis experts confirmed that the extra sub-chromosome "*0*" represents a structural abnormality.

### 4. Conclusion and Future Work

In this paper, we introduce Tokensome, a novel vision-language model based on chromosome tokenization for explainable and cognitive karyotyping. Tokensome elevates the karyotyping method from the conventional visual perception layer to the cognitive decision-making layer. This elevation enables the integration of domain knowledge and cognitive reasoning via knowledge graph and LLM, markedly enhancing model's explainability and facilitating abnormality detection. Our approach demonstrates that embedding higher-order cognitive capabilities into conventional vision models can revolutionize biomedical image analysis and has broad implications for enhancing healthcare diagnostics.

Given that this research is ongoing, we plan to integrate knowledge graphs, Large Language Models (LLMs), and agent-based techniques to facilitate strategic reasoning and domain-aligned optimization. This integration aims to enhance the processes of integrated segmentation, classification, and abnormality detection.

9650-9660. 2021.

[11] Lafferty, John, Andrew McCallum, and Fernando Pereira. "Conditional random fields: Probabilistic models for segmenting and labeling sequence data." In *Icml*, vol. 1, no. 2, p. 3. 2001.

[12] Ziegler, Adrian, and Yuki M. Asano. "Self-supervised learning of object parts for semantic segmentation." In *Proceedings of the IEEE/CVF Conference on Computer Vision and Pattern Recognition*, pp. 14502-14511. 2022.

[13] Han, Jiawei, Jian Pei, and Yiwen Yin. "Mining frequent patterns without candidate generation." *ACM sigmod record* 29, no. 2 (2000): 1-12.

[14] Chen, Delong, Samuel Cahyawijaya, Jianfeng Liu, Baoyuan Wang, and Pascale Fung. "Subobject-level Image Tokenization." *arXiv preprint arXiv:2402.14327* (2024).

[15] Microsoft Research. (2023). Phi-2 Model. Retrieved from: https://huggingface.co/microsoft/phi-2

[16] Lin, Chengchuang, Gansen Zhao, Zhirong Yang, Aihua Yin, Xinming Wang, Li Guo, Hanbiao Chen et al. "Cir-net: Automatic classification of human chromosome based on inception-resnet architecture." *IEEE/ACM Transactions on Computational Biology and Bioinformatics* (2020).

[17] Krizhevsky, Alex, Ilya Sutskever, and Geoffrey E. Hinton. "Imagenet classification with deep convolutional neural networks." *Advances in neural information processing systems* 25 (2012).

[18] Qin, Yulei, Juan Wen, Hao Zheng, Xiaolin Huang, Jie Yang, Ning Song, Yue-Min Zhu, Lingqian Wu, and Guang-Zhong Yang. "Varifocal-net: A chromosome classification approach using deep convolutional networks." *IEEE transactions on medical imaging* 38, no. 11 (2019): 2569-2581.

[19] Zhang, Jiping, Wenjing Hu, Shuyuan Li, Yaofeng Wen, Yong Bao, Hefeng Huang, Chenming Xu, and Dahong Qian. "Chromosome classification and straightening based on an interleaved and multi-task network." *IEEE Journal of Biomedical and Health Informatics* 25, no. 8 (2021): 3240-3251.

[20] Thanh, Vu Duy, Nguyen Huu Hoang Son, Doan Thi Kim Phuong, Luong Thi Lan Anh, Nguyen Thanh Binh Minh, Tran Hoang Tung, Nguyen Hong Thinh, and Luu Manh Ha. "Efficient Type and Polarity Classification of Chromosome Images using CNNs: a Primary Evaluation on Multiple Datasets." In *2022 IEEE Ninth International Conference on Communications and Electronics (ICCE)*, pp. 400-405. IEEE, 2022.

[21] Xia, Chao, Jiyue Wang, Yulei Qin, Juan Wen, Zhaojiang Liu, Ning Song, Lingqian Wu, Bing Chen, Yun Gu, and Jie Yang. "KaryoNet: Chromosome Recognition with End-to-End Combinatorial Optimization Network." *IEEE Transactions on Medical Imaging* (2023).

[22] Yan, Jiahe, Emily Tucci, and Nathaniel Jaffe. "Detection of t (9; 22) chromosome translocation using deep residual neural network." *Journal of Computer and*
10